\PassOptionsToPackage{table}{xcolor}

\documentclass{article}

\usepackage[preprint]{cryptotree}

\usepackage[utf8]{inputenc} 
\usepackage[T1]{fontenc}    
\usepackage{hyperref}       
\usepackage{url}            
\usepackage{booktabs}       
\usepackage{amsfonts}       
\usepackage{nicefrac}       
\usepackage{microtype}      

\usepackage{amsmath}
\usepackage{mathrsfs}  
\usepackage{svg}
\usepackage{bbold}

\usepackage{xcolor}
\usepackage[linesnumbered,ruled,vlined]{algorithm2e}

\SetCommentSty{mycommfont}

\usepackage{tikz}
\usepackage{amsfonts}
\usetikzlibrary{shapes.geometric, arrows, decorations.pathreplacing}

\usepackage{booktabs}
\usepackage{multirow}

\usepackage{natbib}
\setcitestyle{numbers}

\title{Cryptotree : fast and accurate predictions on encrypted structured data}

\author{%
  Daniel Huynh \\
  Intern at Microsoft France\\
  \texttt{daniel.huynh@polytechnique.org} \\
}

\newcommand{\myequation}{\begin{equation}}
\newcommand{\myendequation}{\end{equation}}
\let\[\myequation
\let\]\myendequation

\newcommand{\blue}[1]{\textcolor{blue!70}{#1}}
\newcommand{\green}[1]{\textcolor{green!70}{#1}}
\newcommand{\red}[1]{\textcolor{red!70}{#1}}

\begin{document}

\maketitle

\begin{abstract}
Applying machine learning algorithms to private data, such as financial or medical data, while preserving their confidentiality, is a difficult task. Homomorphic Encryption (HE) is acknowledged for its ability to allow computation on encrypted data, where both the input and output are encrypted, which therefore enables secure inference on private data. Nonetheless, because of the constraints of HE, such as its inability to evaluate non-polynomial functions or to perform arbitrary matrix multiplication efficiently, only inference of linear models seem usable in practice in the HE paradigm so far.

In this paper, we propose \href{https://github.com/dhuynh95/cryptotree}{Cryptotree}, a framework that enables the use of Random Forests (RF), a very powerful learning procedure compared to linear regression, in the context of HE. To this aim, we first convert a regular RF to a Neural RF, then adapt this to fit the HE scheme CKKS, which allows HE operations on real values. Through SIMD operations, we are able to have quick inference and prediction results better than the original RF on encrypted data. 
\end{abstract}

\section{Introduction}

\textbf{Context.} Many areas of Machine Learning have thrived these recent years, nonetheless some domains such as the financial or medical sectors have developed at a slower pace, as they often handle very sensitive data. While some Machine Learning services could greatly benefit society, allowing sensitive information to transit from the client to the server could lead to leaks, as the server could be malicious, or be compromised by a third party and private data could be exposed. Nonetheless, several solutions have emerged such as Secure Multi Party Computation \cite{spdz}, Trusted Execution Environments \cite{enclave}, and Homomorphic Encryption (HE). Each one has its own pros and cons, and throughout this paper we will focus on a solution using the HE paradigm.

HE is an encryption scheme, which allows data owners to encrypt their data, and let a third party perform computations on it, without knowing what is the underlying data. The result of the computations on encrypted data can then be sent back to the data owner, which will be the only one able to decrypt the encrypted result.

More formally, a ring homomorphism $h$ between two rings $R$ and $R'$, follows those two properties : $h(x + y) = h(x) + h(y)$ and $h(x * y) = h(x) * h(y)$. This means that if we have an encryption homomorphism $e$, a decryption homomorphism $d$, such that $d(e(x)) = x$, and a function $f$, which is a composition of additions and multiplications, then we can have the following scenario :t he user encrypts her data $x$ using $e$, and sends $e(x)$ to an untrusted third party. Then the third party performs computations $f$ on the encrypted $e(x)$. Because $e$ is an homomorphism, we have that $f(ex(x)) = e(f(x))$. The third party sends the data back to the user. Finally the user decrypts the output, obtaining then $d(e(f(x))) = f(x)$, without exposing her data directly to the untrusted third party.

HE schemes first started with \cite{rivest}, as RSA provided an homomorphic scheme, but only with homomorphic multiplication. Since then several HE schemes appeared \cite{goldwasser},\cite{elgamal}, \cite{paillier}, \cite{bonneh}, \cite{gentry} and while they allowed more and more operations, they remained too computationally heavy for real use. Recently, more practical schemes have emerged such as BGV \cite{brakerski} and BFV \cite{fan} for integer arithmetic, and CKKS \cite{song} for arithmetic on complex numbers. In this paper we will use CKKS, a leveled homomorphic encryption scheme, which means that addition and multiplication are possible, but a limited number of multiplications is possible. This is due to the fact that noise is injected to the plaintext to hide it, but this noise grows during computation, and above a certain threshold the message cannot be decrypted correctly.

Because of those constraints, the evaluation of a Deep Neural Network (DNN) of arbitrary depth proves to be difficult: we have a limited number of multiplications, matrix multiplication works efficiently only on special cases, and non polynomial functions such as ReLU or sigmoid are difficult to approximate.

However, if we focus on the case of structured data, it is possible to find efficient and expressive models compatible with CKKS. \cite{scornet} gave a very efficient way to model RFs as a DNN with two hidden layers and one output layer, called a Neural Random Forest (NRF). By leveraging this efficient representation of RFs and by tuning it, we show in this paper that one can adapt arbitrary RFs into Homomorphic Random Forests (HRF) which can do quick inference, and have similar performances, if not superior, to the original RF. Thus we show that efficient and expressive models can be used on encrypted data, therefore paving the way for more privacy friendly Machine Learning.

\textbf{Our contribution.} In this paper we show how Random Forests can be modeled efficiently by first converting them to Neural Random Forests, and then implementing their Homomorphic Random Forest counterparts under the CKKS scheme. We show that HRF can leverage the SIMD nature of CKKS to compute the predictions of each homomorphic tree at the same time, resulting in an efficient and expressive model. We then compare the performances of HRF on a real dataset, the Adult Income dataset, and show that HRF performs as well as their neural counterpart, and outperforms the original RF.  

We provide a Python implementation of Homomorphic Random Forest in the Cryptotree library, which uses hooks provided by the TenSEAL library to interact with the C++ SEAL framework developed by Microsoft for HE schemes. Cryptotree provides a high level API so that users who do not have notions of HE can convert their Random Forest models to Neural Random Forests first, fine tune them, then convert these to Homomorphic Random Forests.

\section{Preliminaries}

\subsection{CKKS's scheme} \label{section:ckks}

\tikzstyle{cleartext} = [rectangle, rounded corners,minimum width=3cm, minimum height=1cm, text centered, draw=black, fill=red!30]
\tikzstyle{plaintext} = [rectangle, rounded corners,minimum width=3cm, minimum height=1cm, text centered, draw=black, fill=orange!30]
\tikzstyle{ciphertext} = [rectangle, rounded corners,minimum width=3cm, minimum height=1cm, text centered, draw=black, fill=blue!30]
\tikzstyle{arrow} = [thick,->,>=stealth]

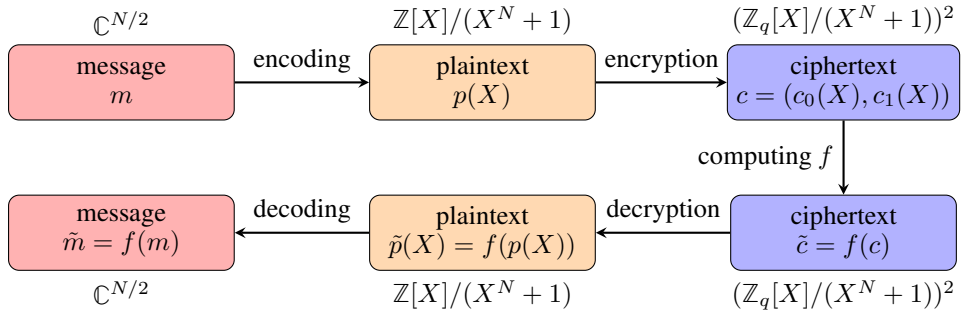
\begin{figure}[htbp]
    \centering
\begin{tikzpicture}[scale=0.8]
\node[cleartext, align = center] (m) at (0,0) {message \\ $m$};
\node[plaintext, align = center] (p) at (6,0) {plaintext \\ $p(X)$};
\node[ciphertext, align = center] (c) at (12,0) {ciphertext \\ $c = (c_0(X), c_1(X))$};

\node[cleartext, align = center] (m2) at (0,-2.5) {message \\ $\tilde{m}=f(m)$};
\node[plaintext, align = center] (p2) at (6,-2.5) {plaintext \\ $\tilde{p}(X) = f(p(X))$};
\node[ciphertext, align = center] (c2) at (12,-2.5) {ciphertext \\ $\tilde{c} = f(c)$};

\draw (m.north) node[above]{$\mathbb{C}^{N/2}$};
\draw (m2.south) node[below]{$\mathbb{C}^{N/2}$};
\draw (p.north) node[above]{$\mathbb{Z}[X]/(X^N + 1)$};
\draw (p2.south) node[below]{$\mathbb{Z}[X]/(X^N + 1)$};
\draw (c.north) node[above]{$(\mathbb{Z}_q[X]/(X^N + 1))^2$};
\draw (c2.south) node[below]{$(\mathbb{Z}_q[X]/(X^N + 1))^2$};

\draw [arrow] (m) -- node[anchor=south] {encoding} (p) ;
\draw [arrow] (p) -- node[anchor=south] {encryption} (c);
\draw [arrow] (c) -- node[anchor=east] {computing $f$} (c2);
\draw [arrow] (c2) -- node[anchor=south] {decryption} (p2);
\draw [arrow] (p2) -- node[anchor=south] {decoding} (m2);
\end{tikzpicture}
    \caption{Overview of CKKS}
    \label{fig:CKKS overview}
\end{figure}

We restate the CKKS \cite{song} scheme here. As this is not the main topic of this paper, we will cover it briefly, underline its shortcomings and we will see how to overcome them when implementing Homomorphic Random Forests. For the interested reader, a more detailed introduction to CKKS is provided in the Appendix.

Figure~\ref{fig:CKKS overview} provides a high level view of CKKS. Let $N$ be a power of two, $M = 2 N$, $\Phi_M(X) = X^N + 1$, the $M$-th cyclotomic polynomial of degree $N$. For efficiency and security reasons, we will work with $\mathcal{R} = \mathbb{Z}[X]/(X^N +1)$ the ring integer of the $M$-th cyclotomic polynomial. We can see that a message $m \in \mathbb{C}^{N/2}$ is first encoded into a plaintext polynomial $p(X) \in \mathcal{R}$ then encrypted using a public key. Once the message is encrypted into $c$ which is a couple of polynomials, CKKS provides several operations that can be performed on it, such as addition, multiplication and rotation. While addition is pretty straightforward, multiplication has the particularity of increasing a lot the noise kept in the ciphertext, therefore to manage it, only a limited number of multiplications are allowed. Rotations are permutations on the slots of a given ciphertext. If we denote by $f$ a function which is a composition of homomorphic operations, then we have that decrypting $\tilde{c} = f(c)$ with the secret key will yield $\tilde{p}(X) = f(p(X))$. Therefore once we decode it, we will get $\tilde{m} = f(m)$.

\textbf{Shortcomings.} While this introduction to CKKS was rather short, there are a few points to take into consideration when applying CKKS to Machine Learning. All inputs are represented as a vector $z \in \mathbb{C}^{N/2}$. If the actual dimension of the input is $k < N/2$, then it will be padded with zeros. This could be inefficient if operations are done on vectors of dimension $k \ll N/2$.

Moreover, only polynomial functions can be computed using the operations allowed in CKKS. While it is possible to approximate non polynomial functions, for instance using Chebyshev polynomials, only low degree polynomials can be evaluated, as CKKS is a leveled scheme, therefore a limited number of multiplications are possible. Moreover, a polynomial interpolation is viable only on a predefined domain, usually $D = [-1,1]$ for Chebyshev interpolation. 

Besides, additions and multiplications are done element-wise on the encrypted inputs. This means that it becomes non trivial to perform simple linear operations such as summing all the coordinates of a vector, or doing a matrix multiplication between a plaintext matrix and a ciphertext vector.

We see that the last two points, combined with the leveled aspect of CKKS, where only a limited number of multiplications can be performed, make the evaluation of arbitrary DNN difficult as we would want to perform any kind of matrix multiplication, and use non polynomial activation functions. 

\textbf{Linear operations.} Nonetheless \cite{halevi} provides a way to do square matrix multiplication by using the diagonals of the matrix. Let $n = N/2$ denote the number of slots available, $A \in \mathbb{R}^{n \times n}$ and $z \in \mathbb{R}^n$, then we have $A . z = \sum_{l=0}^{n-1} \text{diag}(A,l) \odot \text{Rotation}(z, l)$.

where $.$ denotes the matrix multiplication, $\odot$ denotes the coefficient wise vector multiplication, $\text{diag}(A,l) = (A_{0,l}, A_{1,l+1}, ...,A_{n -l -1,n - 1}, A_{n -l, 0},..., A_{n -1, l -1})$ is the $l$-th diagonal of $A$, and $\text{Rotation}(z, l) = (z_{l+1},z_{l+2},...,z_1,z_2,...,z_l)$ the rotation of $z$ by shifting $l$ slots to the left. 

In practice in CKKS when doing matrix multiplication with a size $k < n$, the input vector will be padded with zeros to size $n$, and the same will happen to the diagonals. Therefore when doing the rotation, the first elements will be sent to the end, and zeros will come up and give wrong results. For instance, if we have $k=3$, and $z= (z_1,z_2,z_3,0,\dots,0)$, when we do a rotation of one slot we get $(z_2,z_3,0,\dots,0,z_1)$ therefore when we multiply the rotated input vector with the padded diagonal we will not get the correct result. A solution to this would be to first replicate the first $k-1$ coordinates of $z$, yielding $\tilde{z} = (z_1,\dots,z_k,z_1,\dots,z_{k-1},0,\dots,0)$. Therefore each rotation from $0$ to $k-1$ will output the correct vector to be multiplied with the padded diagonal. Here is an example on vectors of size $k=3$ padded to $n$ : 

$
    \begin{pmatrix}
    \green{a_{1,1}} \\
    \green{a_{2,2}} \\
    \green{a_{3,3}}\\
    0 \\
    \dots \\
    0 \\
    \end{pmatrix}
    \odot
    \begin{pmatrix}
    x \\
    y \\
    z\\
    x \\
    y\\
    \dots \\
    \end{pmatrix}
    + 
    \begin{pmatrix}
    \red{a_{1,2}} \\
    \red{a_{2,3}} \\
    \red{a_{3,1}}\\
    0 \\
    \dots \\
    0 \\
    \end{pmatrix}
    \odot
    \begin{pmatrix}
    y \\
    z \\
    x\\
    y \\
    0 \\
    \dots \\
    \end{pmatrix}
    + 
    \begin{pmatrix}
    \blue{a_{1,3}} \\
    \blue{a_{2,1}} \\
    \blue{a_{3,2}}\\
    0 \\
    \dots \\
    0 \\
    \end{pmatrix}
    \odot
    \begin{pmatrix}
    z \\
    x \\
    y\\
    0\\
    \dots \\
    0\\
    \end{pmatrix}
    = 
    \begin{pmatrix}
    \green{a_{1,1}} x + \red{a_{1,2}} y + \blue{a_{1,3}} z \\
    \blue{a_{2,1}} x + \green{a_{2,2}} y + \red{a_{2,3}} z \\
    \red{a_{3,1}} x + \blue{a_{3,2}} y + \green{a_{3,3}} z \\
    0 \\
    \dots \\
    0 \\
    \end{pmatrix}
$

We generalize this to do $L \geq 1$ matrix multiplications at the same time on $L$ different inputs using only $k$ additions and multiplications (computations are detailed in the Appendix) :  

\begin{algorithm}[H]
\SetAlgoLined
\KwData{$W^{(1)},\dots,W^{(L)} \in \mathbb{R}^{k \times k}, \tilde{z} = (\tilde{z}^{(1)} \ | \ \dots \ | \ \tilde{z}^{(L)} \ | \ 0, \dots, 0) \in \mathbb{R}^{n}$}
\KwResult{$(W_1 . z_1,0 \dots, 0 | \ \dots \ | \ W_L . z_1,0 \dots, 0   \ | \ 0, \dots, 0) \in \mathbb{R}^n$}
$s \leftarrow 0 \in \mathbb{R}^n$\;
\For{$i = 1, \dots, k$}
{
    \For{$l = 1,\dots, L$}
    {
        $D_k^{(l)} \leftarrow \text{diag}(W^{(l)},i) \in \mathbb{R}^k$\ \tcp*{First we extract the diagonal}
        $D_k^{(l)} \leftarrow (D_k^{(l)} | \ 0, \dots, 0) \in \mathbb{R}^{2 k - 1}$\ \tcp*{Then we pad it}
    }
    $D_k \leftarrow (D_k^{(1)} \ | \ \dots \ | \ D_k^{(L)} \ | \ 0,\dots,0) \in \mathbb{R}^n$\ \tcp*{We concatenate them}
    $s \leftarrow s + D_k \odot \text{Rotation}(\tilde{z}, i)$\;
}

 \KwRet{$s$}
 
 \caption{\texttt{PackedMatrixMultiplication}}
 \label{algo:matrix}
\end{algorithm}
This algorithm is quite interesting as we can perform $L$ matrix multiplications of size $k \times k$ at the constant cost of $k$ multiplications and additions as long as we have that $L ( 2 k -1) \leq n$. This particularly fits the context of RFs as $k$ would be the depth of the trees, which will is low because we use shallow trees, and $L$ is high because we have many trees. 

Sum reduction of a vector of size $k \leq n$ can also be performed in a similar fashion. Therefore by combining element-wise multiplication and sum reduction, we can implement dot product in HE (more explanations can be found in the Appendix) :

\begin{algorithm}[H]
\SetAlgoLined
\KwData{$x,y \in \mathbb{R}^{k}$}
\KwResult{$\langle x,y \rangle = \sum_{i=1}^k x_i y_i$}
    $N \leftarrow \lceil \text{log}_2(k) \rceil$\ \tcp*{We compute the number of steps}
    $z \leftarrow x \odot y$\;
    $s \leftarrow$ z\;
 \For{$i = 0, \dots, N - 1$}{
    $t \leftarrow \text{Rotate}(s, 2^i)$\ \tcp*{We rotate and add previous result}
    $s \leftarrow s + t$\;
    $t \leftarrow s$\;
 }
 \KwRet{$s_1$}
 \caption{\texttt{DotProduct}}
 \label{algo:dot}
\end{algorithm}

\subsection{Neural Random Forests} \label{section:nrf}

\begin{figure}[htbp]
    \centering
\begin{tabular}{lcr}
\begin{tikzpicture}[scale=0.6]
 \tikzstyle{neuron}=[circle,  draw=black,
    fill=white, minimum size=12pt,  inner sep=0pt,  font=\sffamily]
     \tikzstyle{leaf neuron}=[rectangle,  draw=black,
    fill=white, minimum size=12pt,  inner sep=0pt,  font=\sffamily]
 \tikzstyle{branch neuron}=[neuron, draw=green, very thick, fill = white, minimum size = 16pt, inner sep = 2pt]
   
    \tikzstyle{special leaf neuron} = [rectangle, draw=black, very thick, fill = red, minimum size = 12pt, inner sep = 2pt]

\draw[very thick] (0,0) rectangle (6,6);
\draw[blue, very thick] (2.5,0) -- (2.5, 6);
\draw (3.5,0) -- (3.5, 6);
\draw[blue, very thick] (0,2) -- (2.5,2);
\draw[blue, very thick] (1.5,2) -- (1.5,6);
\draw (3.5,4) -- (6, 4);

 \node[branch neuron] (I-1) at (2.5,5.5) {$0$};
 \node[branch neuron] (I-2) at (0.75,2) {$1$};
 \node[branch neuron] (I-3) at (1.5,3) {$3$};
 \node[neuron] (I-4) at (3.5,2) {$6$};
 \node[neuron] (I-5) at (5.5,4) {$8$};

 \node[leaf neuron] (II-1) at (1.25,1) {$2$};
 \node[leaf neuron] (II-3) at (2.05,4.25) {$5$};
 \node[leaf neuron] (II-4) at (3,3) {$7$};
 \node[leaf neuron] (II-5) at (4.75,2) {$9$};
 \node[leaf neuron] (II-6) at (4.75,5) {$10$};

 \node[special leaf neuron] (II-1) at (0.75,4.25) {$4$};
\end{tikzpicture}
& 
\hspace{0.5cm}
&
   \begin{tikzpicture}[->,>=stealth',level/.style={sibling distance = 4cm/#1,
  level distance = 1.5cm}, scale = 0.8]

\node {0}
    child{ node  {1}
            child{ node  {2}
            }   
            child{ node  {3}
                            child{ node  {4}}
                            child{ node  {5}}    }                           
    }
    child{ node  {6}
            child{ node  {7} }
            child{ node  {8}
                            child{ node  {9}}
                            child{ node  {10}}
            }
        };
\end{tikzpicture}
\end{tabular}

\def\layersep{2.5cm}
\def\HLOneN{5}
\def\HLTwoN{6}
\def\nFeats{2}
\vspace{0.7cm}

\begin{tikzpicture}[shorten >=0pt,-,draw=black, node distance=\layersep, scale = 0.65]
    \tikzstyle{every pin edge}=[-, draw = black!30]
    \tikzstyle{neuron}=[circle,  draw=black,
    fill=white, minimum size=14pt,  inner sep=0pt,  font=\sffamily]

    \tikzstyle{input neuron}=[neuron];
    \tikzstyle{branch neuron}=[neuron, draw=green, very thick, fill = white, minimum size = 16pt, inner sep = 2pt]
    \tikzstyle{invisible neuron}=[circle, draw = white]
   
    \tikzstyle{leaf neuron}=[rectangle,  draw=black,
    fill=white, minimum size=14pt,  inner sep=0pt,  font=\sffamily]

    \tikzstyle{special leaf neuron} = [rectangle, draw=black, very thick, fill = red, minimum size = 16pt, inner sep = 2pt]
    \tikzstyle{output neuron}=[neuron];
    \tikzstyle{hidden neuron}=[neuron];
    \tikzstyle{annot} = [text width=8em, text centered, font = \footnotesize]

    \foreach \name / \y in {1,...,\nFeats}
        \node[input neuron] (I-\name) at (0,-2-\y) {$x_{\y}$};



             \foreach \name / \y in {1}
        \path[yshift=0.5cm]
         node[branch neuron] (HL1-\name) at (\layersep,-\y cm-0.5 cm) { 0};
        
         \foreach \name / \y in {2}
        \path[yshift=0.5cm]
         node[branch neuron] (HL1-\name) at (\layersep,-\y cm-0.5 cm) { 1};

%

    \foreach \name / \y in {3}
        \path[yshift=0.5cm]
           node[branch neuron] (HL1-\name) at (\layersep,-\y cm -0.5 cm) {3};

    \foreach \name / \y in {4}
        \path[yshift=0.5cm]
           node[hidden neuron] (HL1-\name) at (\layersep,-\y cm -0.5 cm) {6};
    \foreach \name / \y in {5}
        \path[yshift=0.5cm]
           node[hidden neuron] (HL1-\name) at (\layersep,-\y cm -0.5 cm) {8};


  \foreach \name / \y in {1}
        \path[yshift=0.5cm]
            node[leaf neuron] (HL2-\name) at (2*\layersep,-\y cm) {2};
           
   
    \foreach \name / \y in {2}
        \path[yshift=0.5cm]
            node[special leaf neuron] (HL2-\name) at (2*\layersep,-\y cm) {4};
    
    \foreach \name / \y in {3}
        \path[yshift=0.5cm]
            node[leaf neuron] (HL2-\name) at (2*\layersep,-\y cm) {5};
 \foreach \name / \y in {4}
        \path[yshift=0.5cm]
            node[leaf neuron] (HL2-\name) at (2*\layersep,-\y cm) {7};
 \foreach \name / \y in {5}
        \path[yshift=0.5cm]
            node[leaf neuron] (HL2-\name) at (2*\layersep,-\y cm) {9};
 \foreach \name / \y in {6}
        \path[yshift=0.5cm]
            node[leaf neuron] (HL2-\name) at (2*\layersep,-\y cm) {10};

     \node[output neuron,pin={[pin edge={-}]right:}] (O1) at (3*\layersep,-3.5 cm)  {};

  
    \path (I-1) edge (HL1-1)[black, thick];
    \path (I-1) edge (HL1-3)[black, thick];
    \path (I-1) edge (HL1-4)[black!40];
    \path (I-2) edge (HL1-2)[black, thick];
    \path (I-2) edge (HL1-5)[black!40];



\foreach \source in {1,2}
     \path (HL1-\source) edge (HL2-1)[black!40];
\foreach \source in {1,2,3}
     \path (HL1-\source) edge (HL2-3)[black!40];
\foreach \source in {1,4}
     \path (HL1-\source) edge (HL2-4)[black!40];
\foreach \source in {1,4,5}
     \path (HL1-\source) edge (HL2-5)[black!40];
\foreach \source in {1,4,5}
     \path (HL1-\source) edge (HL2-6)[black!40];

   \foreach \source in {1,2,3}
     \path (HL1-\source) edge (HL2-2)[blue, thick];

    \foreach \source in {1,...,\HLTwoN}
        \path (HL2-\source) edge (O1)[black!40];

    \path (HL2-2) edge (O1)[red!90, thick];

    \node[annot,above of=HL1-1, node distance=1cm] (hl1) {$\mathscr{H}$};
    \node[annot,left of=hl1, node distance=2cm] {Input layer};
    \node[annot,right of=hl1, node distance=2cm](hl2){ $\mathscr{L}$};
    \node[annot,right of=hl2, node distance=2cm](hl2){Output layer};

\end{tikzpicture}

    \caption{An example of regression tree (top) and the corresponding neural network (down) from \cite{scornet}}
    \label{fig:nrf}
\end{figure}
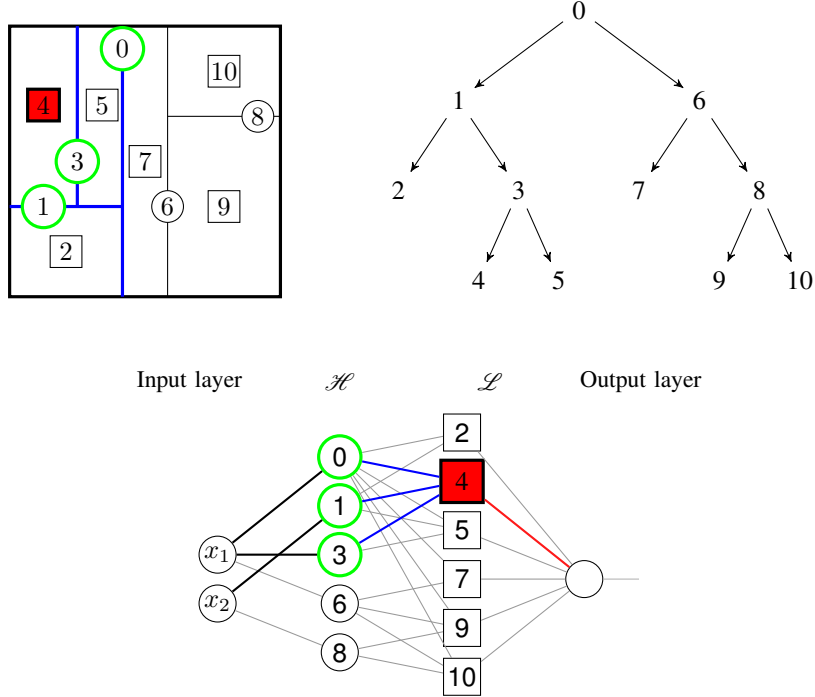

In this section we will see how RFs can be modeled using DNNs. This will serve as a basis for us, as Neural Random Forests have a nice structure, which will enable us to implement them efficiently in CKKS. Using trees to initialize DNNs has been already used in the past, as in \cite{sethi}, \cite{brent} or \cite{lugosi}. We will focus on the recent work of \cite{scornet}, which models RFs efficiently using a DNN with two hidden layers and one output layer. 

We can see on Figure~\ref{fig:nrf} how a DNN can simulate a decision tree. Imagine that an observation belongs to the leaf 4. Then based on the decision tree, this means that it started at the root, the node 0, then went left, went right of the node 1, and finally went left of the node 3. The idea of Neural Random Forests is to first perform all comparisons at the same time at the first layer, then determine which leaf the observation belongs to. Each neuron of the second layer represents a leaf, and only one neuron of the second layer will be activated. The neuron being activated will be the one representing the leaf where the observation lies, and this is done by using the computed comparisons of the first layer to determine exactly where the observation is. Finally, once the observation has been located, the final output layer will simply output the mean of this leaf for regression, or the distribution of this leaf for classification.

We will now see more formally how Neural Random Forests are modeled. Let $d \in \mathbb{N}^{\star}$, $X = [0,1]^d$ our normalized input space, and $Y = \mathbb{R}$ the output space for regression, or $Y = \Delta(\mathbb{R}^C)$ the probability simplex for classification. We assume we are given dataset $\mathcal{D}_n = ((X_1,Y_1),...,(X_n,Y_n))$, with $n \geq 2$. Let $T$ be a binary decision tree. We denote by $K$ the number of leaves, i.e. terminal nodes in the tree. Notice that if a binary decision tree has $K$ leaves, it has necessarily $K - 1$ internal nodes, which represent the comparisons.

Let $x \in X$ be an observation, $H = (H_1, ..., H_{K-1})$ be the collection of hyperplanes used in the construction of $T$. For $k \in [1 \dots K- 1]$ we have that $H_k = \{ x \in X : h_k(x) = 0  \}$, with $h_k(x) = x_{\tau(k)} - t_k$, $\tau : [1 \dots K- 1] \rightarrow [1 \dots D]$ where $\tau(k)$ is the index of the variable used in comparison $k$, and $t_k \in [0,1]$ is the threshold value of the comparison $k$. Therefore the first linear layer simply applies comparisons, and apply a non linearity $\phi(x) = 2 \mathbb{1}_{x \geq 0} - 1$ which will send to $+1$ if the variable was above the threshold or $-1$ otherwise. Thus the $K - 1$ outputs of the first layer are : 
\[ \label{eq:comparison}
u_k(x) = \phi(x_{\tau(k)} - t_k), k \in [1 \dots K- 1] 
\]
The second hidden layer will take the $K - 1$ output bits $\pm 1$, and output $K$ bits, with 1 bit +1 at the index of the leaf where the observation belongs to, and -1 everywhere else. To do so, let us denote by $k' \in [1 \dots K]$ the index of a given leaf, and define $v_{k'}(x)$ one output of the second layer. We will use non zero weights for the previous $u_k(x)$ if and only if the comparison $k$ was used to reach the leaf $k'$. The connection has weight +1 if, in that path, the split by $H_k$ is from a node to a right child, and -1 otherwise. Then we have that : 
\[ \label{eq:match}
v_{k'}(x) = \phi(\sum_{k \rightarrow k'} V_{k,k'} u_k(x) + b_{k'}), k' \in [1 \dots K] 
\]
where notation $k \rightarrow k'$ means that $k$ is connected to $k'$ and $V_{k,k'} = \pm 1$ is the corresponding weight. Here $b_{k'} = - l(k') + \frac{1}{2}$, with $l(k')$ the length of the path from the root to the leaf $k'$. One important thing to note when we will implement this later in CKKS is that : 
\[ \label{eq:match_bounds}
-2 l(k') + \frac{1}{2} \leq \sum_{k \rightarrow k'} V_{k,k'} u_k(x) + b_{k'} \leq \frac{1}{2}
\]
More specifically, we have that $\sum_{k \rightarrow k'} V_{k,k'}(x) u_k(x) + b_{k'}$ is positive if and only if $x$ belongs to the leaf $k'$. Therefore the output of the second layer exactly outputs +1 at the leaf the observation belongs to, and -1 otherwise.

Finally the output of the tree, denoted $T(x)$ is :
\[ 
T(x) = \sum_{k' = 1}^{K} W_{k'} v_{k'}(x) + \beta 
\]
with $\beta = \frac{1}{2n} \sum_{i=1}^{n} Y_i$, $W_{k'} = \frac{1}{2 |L_{k'}|} \sum_{i \in L_{k'}} Y_i$ and $L_{k'}$ is the indexes of the observations which belong to the leaf $k'$. Note that if we are doing regression we will simply output the mean of the target in the leaf, and for classification we will output the distribution of classes within that leaf.

While using $\phi(x) = 2 \mathbb{1}_{x \geq 0} - 1$ provides exactly the same values as the original tree, this activation is non differentiable. By using $\phi_a(x) = \text{tanh}(a x)$, with $a > 0$ a hyper-parameter used as a dilatation factor, it becomes possible to fine tune the DNN.

To generalize from a single tree to a RF with $L \geq 1$ trees $T^{(1)} \dots T^{(L)}$, we can simply compute all predictions in parallel, and given weights $\alpha_l$, take a weighted sum of the output of each tree to obtain the prediction $\hat{y}$ of the RF : 
\[ \label{eq:forest}
\hat{y} = \sum_{l=1}^{L} \alpha_l T^{(l)}(x) 
\]  
\section{Homomorphic Random Forest} \label{section:hrf}

In this section, we will present how Neural Random Forests can be adapted in CKKS. We saw in Section~\ref{section:nrf} that evaluating a NRF can be done using three equations : (\ref{eq:comparison}), (\ref{eq:match}), and (\ref{eq:forest}). Based on the observation that the first layer is a particular type of sparse matrix multiplication where each row selects only one variable, and that the last layer can be seen as doing a few dot products if $C$ is low, we see that to implement HRFs we only need to be able to do a parallel matrix multiplication for each tree at the second layer. 

Instead of performing a matrix multiplication for (\ref{eq:comparison}), we decide to let the user directly perform the linear operation, which is simply a reshuffling of the variables used in the comparison. We chose to do this because this operation has a high performance cost in CKKS, but can be done directly on unencrypted data without compromising the confidentiality of the model.

Then for the second layer (\ref{eq:match}) we will use Algorithm~\ref{algo:matrix} to perform all matrix multiplications in parallel. Algorithm~\ref{algo:matrix} is particularly well suited for our scenario, because we can evaluate all $L$ trees at the same time, with no additional cost, no matter the number of inputs $d$, or the number of trees $L$ as this operation's complexity only depends on the number of leaves $K$.

Finally for the last layer (\ref{eq:forest}), we will compute the score of each class $c$ by noticing that it can actually be seen as a dot product between $v$ and a vector $W_c$ as $\hat{y}_c = \sum_{l=1}^L \alpha_l T_c^{(l)}(x) = \sum_{l=1}^L \alpha_l (\sum_{k'=1}^K W_{c,k'}^{(l)} v_{k'}^{(l)}(x) + \beta_c^{(l)}) = \langle W_c, v \rangle + \beta_c$ where $W_c$ is the concatenation of the weighted $c$-th rows of each $W^{(l)}$, plus a bias that is a weighted sum of the biases of the $L$ trees for class $c$. We will therefore use Algorithm~\ref{algo:dot} to compute this homomorphically. 

One important detail is that we will use a polynomial approximation $P$ of degree $m$ of the regular activation $\phi_a$ used in NRF in equations (\ref{eq:comparison}) and (\ref{eq:match}). As such, we need to make sure that the output of the first and second linear layers are in $[-1,1]$. While it is easy to check this for (\ref{eq:comparison}), as $x \in [0,1]$ and $t \in [0,1]$, it is non trivial for (\ref{eq:match}). Nonetheless, using (\ref{eq:match_bounds}) we see that if we divide it by $2 l(k')$ the weights and the bias, we have that the output of the linear part of (\ref{eq:match}) will be in $[-1,1]$. As the final output layer does not use an activation, we do not need to check if it is in $[-1,1]$.

We therefore propose Algorithm~\ref{algo:hrf} to evaluate homomorphically a NRF composed of trees $(T^{(l)} = (\tau^{(l)}, t^{(l)}, V^{(l)}, b^{(l)}, W^{(l)}, \beta^{(l)}))$, on an input $x \in X$ with a polynomial activation function $P \in \mathbb{R}[X]$. For convenience, we suppose that all trees have been padded to the same number of leaves $K$, the $V^{(l)}$ have been padded to square matrices and $V^{(l)}$ and $b^{(l)}$ have been divided by the appropriate factor so that (\ref{eq:match}) is in $[-1,1]$. We provide in Table~\ref{table:complexity} an overview of the complexity of each linear layer in terms of homomorphic operations. Note that in CKKS additions have a complexity of $\mathcal{O}(n)$, while multiplications and rotations have a complexity of $\mathcal{O}(n \text{log}(n))$, with $n= N/2$ the number of slots in CKKS. 

\begin{algorithm}[H]
\SetAlgoLined
\KwData{$(\tau^{(l)}, t^{(l)}, V^{(l)}, b^{(l)}, W^{(l)}, \beta^{(l)}), \ x \in X, \ P \in \mathbb{R}[X]$}
\KwResult{$\hat{y} \in \mathbb{R}^C$}
Client\ \tcp*{The client will prepare the data before sending it to the server}
\For{$l=1,\dots,L$}
{
    $\tilde{x}^{(l)} \leftarrow (x_{\tau} \ | \  0 \ | \ x_{\tau}) \in \mathbb{R}^{2 K - 1}$\ \tcp*{The input of each tree is replicated}
}
$\tilde{x} \leftarrow (\tilde{x}^{(1)} \ | \ \dots \ | \ \tilde{x}^{(L)} \ | \ 0,\dots, 0) \in \mathbb{R}^{N/2}$\ \tcp*{All inputs are concatenated}
$\tilde{x} \leftarrow \text{Encrypt}(\tilde{x})$\ \tcp*{After preprocessing the client encrypts the data}
Server\;
\For{$l=1,\dots,L$} 
{
    $\tilde{t}^{(l)} \leftarrow (t_{\tau} \ | \  0 \ | \ t_{\tau})$\ \tcp*{Thresholds are prepared similarly to the inputs}
    $\tilde{b}^{(l)} \leftarrow (b_{\tau} \ | \ 0, \dots, 0)$\ \tcp*{Bias needs only to be padded}    
}
$\tilde{t} \leftarrow (\tilde{t}^{(1)} \ | \ \dots \ | \ \tilde{t}^{(L)} \ | \ 0,\dots, 0)$\;
$\tilde{b} \leftarrow (\tilde{b}^{(1)} \ | \ \dots \ | \ \tilde{b}^{(L)} \ | \ 0,\dots, 0)$\;
$u \leftarrow P( \tilde{x} - \tilde{t})$\ \tcp*{We perform the comparisons}
$v \leftarrow P(\texttt{PackedMatrixMultiplication}(W_1^{(1)},\dots,W_1^{(L)},u) + \tilde{b})$\;
\tcc{We then compute leaf scores in parallel}
\For{$c=1,\dots,C$}
{
    \For{$l=1,\dots,L$} 
    {
        $\tilde{W}_c^{(l)} \leftarrow (\alpha_l W_c^{(l)} | \ 0, \dots, 0)$\;
    }
    $\tilde{W}_c \leftarrow (\tilde{W}_c^{(1)} \ | \ \dots \ | \ \tilde{W}_c^{(L)} \ | \ 0,\dots, 0)$\ \tcp*{We compute the dot product weights}
    $\beta_c \leftarrow \sum_{l=1}^L \alpha_l \beta_c^{(l)}$\;
    $\hat{y}_c \leftarrow \texttt{DotProduct}(\tilde{W}_c ,v) + \beta_c$\ \tcp*{We compute the score of each class}
}
 \KwRet{$\hat{y}$}
 \caption{\texttt{HomomorphicRandomForestEvaluation}}
 \label{algo:hrf}
\end{algorithm}

\begin{table}[]
\centering
\caption{Complexity of each linear layer of HRFs}
\label{table:complexity}
\begin{tabular}{@{}llll@{}}
\toprule
                      & Addition & Multiplication & Rotation \\ \midrule
First linear layer    & 1        & 0              & 0        \\
Second linear layer   & $K$        & $K$              & $K$        \\
Third linear layer    & $C \lceil \text{log}_2 L (2K -1) \rceil$ & $C$ & $C \lceil \text{log}_2 L (2K -1) \rceil$    \\ \bottomrule
\end{tabular}
\end{table}

\section{Experimental results}
We will see how HRFs perform compared to other baseline models on the Adult Income dataset \cite{adultincome}, which contains 48842 observations of people's socio-demographics such as age, education, or employment in order to predict if the person's salary is higher than 50K annually. 

We have kept preprocessing minimal as the goal is not to have the best prediction score, but to see how HRFs compare to other models. Continuous data was normalized to $[0,1]$, and categorical data was label encoded, then normalized to $[0,1]$ as well. Code and further details about the training and evaluation can be found in the Appendix.

\begin{table}[]
\centering
\caption{Results on the Adult Income Dataset}
\label{table:results}
\begin{tabular}{@{}lllll@{}}
\toprule
Model & Accuracy & Precision & Recall & F1
\\ \midrule
Linear & 0.819& 0.432 &	0.724 &	0.541 \\
RF & 0.834 & 0.386 & \textbf{0.876} & 0.536 \\
NRF & \textbf{0.845} & \textbf{0.547} & 0.762 & \textbf{0.637} \\
HRF & 0.842 & 0.491 & 0.796 & 0.607
\\ \bottomrule
\end{tabular}
\end{table}

In order to train a HRF, we first train a regular RF using Scikit-learn. Then we create a Neural Random Forest Pytorch module, based on the trees of the RF. Because we use polynomial activations and not hard comparisons, intermediate results will not be exact, i.e. the output of the first and second layers will not be $\pm 1$. Therefore it is necessary to fine tune the NRF to make it output good predictions when dealing with soft intermediate outputs. Nonetheless, one problem emerges, as we need to make sure that the output of the first and second linear layers are in $[-1,1]$ otherwise when using the approximated activation function it will blow up. Therefore, we only fine tuned the last linear layer, as we do not compute any polynomial after that. 

In Table~\ref{table:results} we compare the results of different models on the validation set. As simple baselines, we added a logistic regression and the original RF. We can see here that the fine tuned NRF actually has better performances than other models. Nonetheless HRF has very similar results, but not exactly the same, as noise is introduced during computation and grows over time. In 97.5\% of the time, the NRF and HRF gave the same results. To mitigate the risks of NRF and HRF giving different outputs, we trained the NRF using Label Smoothing \cite{labelsmoothing}, which allowed to make sure the score of the predicted class is as far as possible from the other scores. As predictions from HRFs are almost identical to NRFs, the interested reader can read more theoretical and practical results in \cite{scornet}.

\section{Related work}

\cite{linearckks} provided explanations on how to train and do inference with logistic regression on encrypted data. On the other hand, DNNs have also been explored in HE, with CryptoNet \cite{cryptonet} which shows how DNN's with convolutions, pooling and dense layers can be implemented in HE. Inference is done by batching 8192 images together. 

Therefore our model seems to be the most relevant for the case of structured data prediction as it is more expressive than logistic regression and more practical than DNNs such as CryptoNet. Even though CryptoNet has good amortized performance when computing on a batch of 8192 samples, it still requires 570 seconds to process a batch on a PC with a single Intel Xeon E5-1620 CPU running at 3.5GHz,with 16GB of RAM using C++. This batching of several inputs suffers from several issues. First of all a user might not care about having her input batched with other inputs and would want the result as fast as possible. Secondly, in practice it is not possible to batch different users' input once they have been encrypted by different public keys, and while it is possible to batch different users' input before they have been encrypted, this assumes that the server saw the data in clear before batching, which defeats the purpose of HE. 

In comparison, HRFs have a much quicker inference time (3s on an Intel Core i7 4600U CPU @ 2.10GHz with 8GB of RAM using Python with C++ bindings) for a single observation, which is much more reasonable. Moreover, several inputs can be handled at the same time using a multi-threaded server. While CryptoNet might have a better amortized performance in theory with batching, we saw that batching is not likely to be used in practice for most scenarios. 

Finally, HRFs do not need much communication compared to SMPC schemes \cite{spdz}, \cite{securenn}, and can be done without the need of a trusted crypto-provider whose role is to hide each party's data, therefore reducing the number of failure nodes.

\section{Conclusion}

In this paper we showed that neuronal approximations of Random Forests can be adapted for in an homomorphic setting. We have seen that this leads to expressive and fast models for inference on encrypted structured data. We will investigate in future work how to fine tune all layers, not just the last one as it is currently done, while preserving the output to be in $[-1,1]$ for correct polynomial evaluations. 

\newpage

\section{Broader impact}

In a context of scandals around data privacy, and new legislation such as GDPR, both companies, institutions and individuals long for more privacy friendly services. While the first fields that would be served from such confidential algorithms are the health and financial sector, nearly every sector would benefit from it. 

We believe that as encrypted protocols such as HTTPS have become the standard, privacy preserving Machine Learning should become the new standard. 

But in order for this revolution to happen, we need both simple, expressive and practical models to make them adopted by a large community. 

While HRFs work only for structured data, they pave the way for practical and accurate inference of Machine Learning models on encrypted user data. While other solutions provide different trade offs between practicality and performance, HRFs strike a good balance between those two notions as it needs little communication as opposed to SMPC, needs no third party, is quick and still can be extremely expressive, as it is a mix of Random Forests and Neural Networks.

Thanks to these strength, its high level API, and the use cases that can be covered, we think that Cryptotree would help accelerate the emergence of more privacy friendly services.

\bibliography{cryptotree}

\end{document}